\documentclass[lettersize,journal]{IEEEtran}
\usepackage{amsmath,amsfonts}
\usepackage{algorithmic}
\usepackage{array}
\usepackage[caption=false,font=normalsize,labelfont=sf,textfont=sf]{subfig}
\usepackage{textcomp}
\usepackage{stfloats}
\usepackage{csquotes}
\usepackage{url}
\usepackage{verbatim}
\usepackage{booktabs}
\usepackage{multirow}
\usepackage{xcolor}
\usepackage{graphicx}
\def\BibTeX{{\rm B\kern-.05em{\sc i\kern-.025em b}\kern-.08em
    T\kern-.1667em\lower.7ex\hbox{E}\kern-.125emX}}
\usepackage{balance}
\begin{document}
\title{Enhancing Document Information Analysis with Multi-Task Pre-training: A Robust Approach for Information Extraction in Visually-Rich Documents}
\author{Tofik~Ali, Partha~Pratim~Roy
\thanks{T. Ali and P.P. Roy are with the Department
of Computer Science and Engineering, Indian Institute of Technology Roorkee, India,
e-mail: tali@cs.iitr.ac.in, e-mail: proy.fcs@iitr.ac.in}
}


\maketitle

\begin{abstract}
This paper introduces a deep learning model tailored for document information analysis, emphasizing document classification, entity relation extraction, and document visual question answering. The proposed model leverages transformer-based models to encode all the information present in a document image, including textual, visual, and layout information. The model is pre-trained and subsequently fine-tuned for various document image analysis tasks. The proposed model incorporates three additional tasks during the pre-training phase, including reading order identification of different layout segments in a document image, layout segments categorization as per PubLayNet, and generation of the text sequence within a given layout segment (text block). The model also incorporates a collective pre-training scheme where losses of all the tasks under consideration, including pre-training and fine-tuning tasks with all datasets, are considered. Additional encoder and decoder blocks are added to the RoBERTa network to generate results for all tasks. The proposed model achieved impressive results across all tasks, with an accuracy of 95.87\% on the RVL-CDIP dataset for document classification, F1 scores of 0.9306, 0.9804, 0.9794, and 0.8742 on the FUNSD, CORD, SROIE, and Kleister-NDA datasets respectively for entity relation extraction, and an ANLS score of 0.8468 on the DocVQA dataset for visual question answering. The results highlight the effectiveness of the proposed model in understanding and interpreting complex document layouts and content, making it a promising tool for document analysis tasks.
\end{abstract}

\begin{IEEEkeywords}
Document Information Analysis, Document Classification, Entity Relation Extraction, Document Visual Question Answering, Deep Learning, Transformer Models, Pre-training Tasks.
\end{IEEEkeywords}

\section{Introduction}
\label{Introduction}
\subsection{Background}
Document information analysis, a pivotal component of Document AI, entails extracting visual information, often termed visual information extraction (VIE). This area of research has garnered significant attention from academic researchers and industry professionals. It primarily focuses on understanding and interpreting visually rich documents (VrDs), such as forms and receipts. These interpretations involve the semantic entities recognition~(SER) and the subsequent relations extraction~(RE). How document layouts are geometrically represented plays a crucial role in this endeavour. Recent advancements have seen pre-training techniques revolutionize the Document AI domain, yielding significant improvements in document comprehension tasks. Such pre-trained models can dissect the layout and pinpoint essential data from diverse documents, ranging from scanned forms to scholarly articles. This capability has notable implications in both industrial applications and academic investigations.

Transformer-based models~\cite{xu2020layoutlm, xu2021layoutlmv2, huang2022layoutlmv3}, rooted in deep learning paradigms, aspire to encapsulate all facets of information present in a document image, encompassing the textual, visual, and layout dimensions. Once this information is encoded within the pre-trained model, it is subsequently fine-tuned for various document image analytical tasks. The prowess of these models is evident in their performance in VIE assignments, especially in SER tasks. However, there's a notable disparity regarding the RE task, which targets identifying relationships between semantic entities within documents. This particular task poses inherent challenges and needs to be more explored.

\subsection{Motivation}
While the aforementioned transformer-based models have exhibited commendable results in many areas, recent findings indicate a mismatch in relationships inferred by these models when tasked with RE within document images~\cite{luo_geolayoutlm_2023}. The challenges with RE stem from the objective gap between the pre-training and fine-tuning phases of these models. Although some works have proposed specific pre-training tasks to bridge this gap, results on datasets like FUNSD have not consistently demonstrated expected improvements. This implies potential hidden issues and emphasizes the importance of more refined layout representation in pre-trained models.

Moreover, the realization that current models might not efficiently comprehend the geometric layout nuances of documents makes it clear that there's an urgent need to investigate more discriminative methods of understanding document layouts. Incorporating explicit modelling of geometric relationships during the pre-training phase emerges as a promising solution to these challenges.

\subsection{Objectives}
The primary objective of this paper is to present a model tailored to address the challenges mentioned earlier. By incorporating innovative pre-training tasks and methodologies, the proposed model significantly enhances RE task performance. Emphasis has been placed on discerning the reading sequence of layout segments within document images, categorizing these layout segments according to standards such as PubLayNet, and generating text sequences within identified layout segments or text blocks.

\subsection{Contributions}
The salient contributions of this research are threefold:

\begin{itemize}
\item A novel collective pre-training strategy is introduced. This strategy incorporates the losses from all tasks into each parameter update step during pre-training. By leveraging datasets like RVL-CDIP~\cite{harley2015evaluation}, FUNSD~\cite{jaume2019funsd}, CORD~\cite{park2019cord}, SROIE~\cite{SROIE}, Kleister-NDA~\cite{Kleister-NDA}, DocVQA~\cite{mathew2021docvqa}, and PubLayNet~\cite{zhong2019publaynet}, we not only improve the quality of the pre-training data but also manage to reduce its size from 11 million~(IIT-CDIP) to just one million, resulting in less pre-training overall computation.

\item Integration of three distinct tasks during the model's pre-training phase. These tasks involve: 1) categorizing layout segments as per PubLayNet, 2) determining the reading order amongst varied layout segments in document images, and 3) producing text sequences within specified layout segments or text blocks.

\item Enhancement of the RoBERTa~\cite{liu_roberta_2019} network architecture by integrating additional encoder and decoder blocks. These blocks are pivotal for generating results across all tasks while minimizing disruption to the core RoBERTa network. The proposed model's pre-training phase fully utilizes these blocks, updating the relevant parameters in line with their respective losses.
\end{itemize}

Building on the findings of previous studies~\cite{huang2022layoutlmv3, luo_geolayoutlm_2023}, this paper introduces the contributions as mentioned above to advance the field of Document AI, particularly in relation extraction tasks.

\section{Related Work}
\label{Related Work}
\subsection{Progress in Multimodal Self-Supervised Pre-training for Document Intelligence}
Document Intelligence, a specialized segment of artificial intelligence aimed at distilling valuable knowledge from textual documents, has immensely benefited from multimodal self-supervised pre-training methodologies. These approaches synthesize text, visuals, and layout data to deepen the comprehension of document frameworks.

The pioneering efforts in this domain can be credited to LayoutLM~\cite{xu2020layoutlm, xu2021layoutlmv2, huang2022layoutlmv3} and its further iterations. These models were trailblazers in amalgamating layout representation by assimilating spatial coordinates of text \cite{xu2020layoutlm, li2021structurallm, hong2022bros, lee2022formnet}. These methodologies hinge on the premise that the strategic positioning of text and visuals in a document offers indispensable insights into its content and structure, facilitating processes such as entity recognition and document classification.

Recognizing the potential of visual data, an innovative fusion of convolutional neural networks (CNNs) and transformer attention mechanisms was explored \cite{vaswani2017attention}. Initial endeavours sought to harness CNN grid features \cite{xu2021layoutlmv2, appalaraju2021docformer} or capitalize on region features via object detection tools \cite{xu2020layoutlm, gu2021unidoc, li2021selfdoc, powalski2021going}. Nevertheless, these ventures often encountered computational hurdles or required specific regional directives.

The evolution in natural image analytics, especially in the domain of vision-and-language pre-training (VLP), paved the way for a paradigm shift from region features to grid features~\cite{chen2020uniter, su2019vl, tan2019lxmert, huang2021seeing}. This adjustment sought to counteract the limitations associated with preset object categories and region-centric guidance. Vision Transformers (ViT)~\cite{dosovitskiy2020image} championed the concept of leveraging image embeddings, sidestepping the traditional CNNs. This concept introduced a fresh array of VLP techniques. Despite several methodologies adhering to distinct self-attention mechanisms, ViLT~\cite{kim2021vilt} marked a turning point by deploying a consolidated linear layer for visual feature extraction, which translated to a notable reduction in model size and processing duration. Building on this foundation, LayoutLMv3~\cite{huang2022layoutlmv3} was conceptualized, standing out as the avant-garde multimodal model in Document AI, uniquely operating without the dependency on CNNs.

\subsection{Evolution through Reconstructive Pre-Training Paradigms}
The landscape of representation learning has witnessed transformative changes owing to reconstructive pre-training paradigms. These paradigms endeavour to encapsulate the intrinsic nature of data, thereby enhancing model adaptability across a spectrum of tasks. In the realm of natural language processing (NLP), "masked language modelling" (MLM) marked a pivotal shift. This strategy, propelled by BERT~\cite{kenton2019bert}, aimed at bidirectional learning by obscuring certain input words and prompting the model to deduce them from the surrounding context. This approach set new benchmarks for several language comprehension tasks as a foundation for subsequent advancements \cite{huang2022layoutlmv3}.

In the domain of computer vision (CV), a parallel approach, termed Masked Image Modeling (MIM), surfaced. This approach took cues from NLP's MLM, intending to deduce concealed image portions based on the observable context. Vision Transformer (ViT)~\cite{dosovitskiy2020image} epitomized this by estimating the average shade of concealed segments, augmenting its prowess on tasks like ImageNet categorization. BEiT~\cite{bao2021beit} further expanded on this concept, zeroing in on visual token revival, and yielded notable outcomes in visual categorization and semantic demarcation. Shifting the focus to documents, Document Transformer (DiT) harnessed these principles for document image layout examination \cite{li2022dit}.

Drawing inspiration from the triumphant trajectories of MLM and MIM in NLP and CV, respectively, scholars blending vision and language embarked on probing reconstructive paradigms tailored for multimodal representation. While the foundational idea of obscuration and prediction persisted, the actual masking modalities were refined according to image embedding specifics. This evolution led to three distinct MIM variants: regional obscuration modelling (MRM), grid-based obscuration modelling (MGM), and segment-based obscuration modelling (MPM). Among these, MRM demonstrated prowess in regaining original regional attributes or discerning labels of obscured regions. In contrast, MGM adeptly deduced visual lexicon linkages for obscured grids. When it came to segment-level representations, platforms like ViLT~\cite{kim2021vilt} and METER forged paths reminiscent of ViT~\cite{dosovitskiy2020image} and BEiT~\cite{bao2021beit}, underscoring the technique's promise, even if they sometimes faced hurdles in certain assignments.

The advent of LayoutLMv3~\cite{huang2022layoutlmv3} signified a noteworthy evolution. Inspired by ViLT~\cite{kim2021vilt}, it emerged as Document AI's pioneer in harnessing image embeddings sans the use of CNNs. This shift not only optimized computations but also enhanced the nuances of document layout portrayals. Moreover, it ratified the potential of MIM in handling linear segment image embeddings, casting light on promising avenues for further exploration \cite{huang2022layoutlmv3, luo_geolayoutlm_2023}.

To sum up, reconstructive pre-training paradigms have charted a transformative course in representation learning across disciplines, such as audio signal processing, medical imaging, social network analysis, and e-commerce product categorization. Bridging the capabilities of NLP and CV, they have paved the way for innovative models adept at comprehending and managing multimodal content, contributing substantially to intricate endeavours such as document layout examination \cite{huang2022layoutlmv3, luo_geolayoutlm_2023}. 

\subsection{Extraction of Visual Information from Document Images}
Deriving Visual Information (VDI) from document images is pivotal for automating the comprehension of documents. This mainly addresses tasks such as Semantic Entity Identification (SEI) and Entity Relationship Analysis (ERA)\cite{hong_bros_2022, jaume2019funsd, li2021structext}. Earlier VDI approaches majorly employed Graph Neural Networks (GNN)\cite{luo2020merge, yu2021pick}, focusing on acquiring features that represent text and layout components for immediate VDI tasks.

The introduction of pre-training approaches has profoundly expanded the scope of document comprehension. Researchers have ingeniously proposed several pre-training tasks aimed at enhancing textual and visual features while ensuring a solid alignment for a sturdy multimodal document interpretation~\cite{li2021structext, luo2022bi, wang2022lilt, xu2020layoutlm, xu2021layoutlmv2}. While significant strides have been made in SEI, ERA is still a less explored domain~\cite{hong_bros_2022, li2021structext, zhang2021entity}. Significantly, BROS~\cite{hong_bros_2022} integrated spatial text positional data into the BERT architecture~\cite{kenton2019bert}, amplifying layout interpretation. This paper emphasizes harnessing pre-training approaches to garner enhanced features for document examination.

An essential aspect of VDI is the spatial details within document structures, which often act as innate indicators for document layout interpretation. As an illustration, Liu et al.\cite{liu2019graph} deployed 2D spatial positions with GNN, while GraphNEMR\cite{luo2020merge} combined geometry proximities and distance metrics for SEI. Innovations such as SPADE~\cite{hwang2021spatial} revamped the self-attention mechanism, embedding spatial vectors that incorporate coordinates, distances, and angular data. StrucText~\cite{li2021structext} introduced a task during pre-training to determine the geometric orientation between text blocks. Yet, there's an evident shortfall as most techniques focus solely on pair-level spatial connections. Our approach seeks to expand these connections to multiple pairs and groups, ensuring a comprehensive investigation.

Furthermore, as pinpointed by~\cite{luo_geolayoutlm_2023}, mismatches between the pre-training and task-specific tuning stages often pose intricate issues. Many recent endeavors~\cite{gururangan2020don, han2021adaptive, howard2018universal, hu2022p3, liu2023pre} are directed at bridging this inconsistency. For example, Hu et al.\cite{hu2022p3} discerned gaps in the training design and task knowledge, ingeniously synchronizing the subsequent ranking task more seamlessly with a pre-training design. The emergence of prompt-centric models has forged a route, allowing models to adapt across diverse contexts by converting specific tasks to associated prompts, aligning seamlessly with the pre-training design\cite{liu2023pre}. Drawing from these advancements, our research infuses spatial tasks during the pre-training stage, ensuring superior incorporation of spatial understanding. This bolsters the model's adaptability, especially in relation extraction, by capitalizing on extensive pre-training resources.

\subsection{Aligning Vision and Language in Multimodal Frameworks}
Integrating visual and textual information in multimodal models poses distinct challenges, primarily regarding the formation of a cohesive understanding from both modalities. Pioneering work in this arena has spanned both overarching and intricate mechanisms for vision-language (VL) alignment.

At the broader scale, VL alignment seeks to comprehend overarching relationships between visual elements and their associated text. One key approach in this realm has been matching images with their corresponding textual descriptions, serving to deepen our grasp on the synergy between visual and textual contexts \cite{huang2022layoutlmv3}. Such general alignments form the foundational understanding for more refined alignment techniques.

Digging deeper, intricate VL alignment targets the intricate mappings between individual textual units and specific regions in images. Document images, distinct from typical pictures, often present clear correlations between their text and visual sections. UNITER~\cite{chen2020uniter}, for instance, implements a word-region correlation mechanism via optimal transports, calculating the least resource-intensive way to relate image-based contextual representations with individual words \cite{huang2022layoutlmv3}. ViLT~\cite{kim2021vilt} then broadens this goal, aiming for patch-level image contextualizations.

Recognizing the idiosyncrasies of document images, UDoc leveraged techniques like contrastive learning and distilled similarities to sharpen the correlation of text and image sectors within the same segments \cite{huang2022layoutlmv3}. In a related vein, LayoutLMv2 honed this by introducing a masking technique on particular textual sections in images, prompting the model to anticipate the hidden textual components.

A recurring tactic across these methods has been the application of masking, exemplified by practices in Masked Image Modeling (MIM), which helps pinpoint both correlated and non-correlated pairs \cite{huang2022layoutlmv3}. This underlines the importance of flexibility in VL alignment strategies. Especially concerning document images, tailoring alignment methods to the dataset's particularities is imperative.

Straying from models that predominantly focus on general images, specialized architectures such as Luo's GeoLayoutLM address challenges specific to document images \cite{huang2022layoutlmv3}. Given the inherent structured relationships in document images, they open doors to pioneering alignment techniques like contrastive learning. This fosters a more seamless connection between the visual and textual aspects.

Conclusively, the interconnection of vision and text in multimodal models is a thriving domain. The continual emergence of innovative methodologies and richer data sources indicates that alignment methods will evolve in response to the multifaceted requirements of diverse visual-textual contexts.

\subsection{Bridging the Gap between Pre-training and Fine-tuning}
Achieving seamless progression from the pre-training phase to the fine-tuning phase is fundamental in today's deep learning landscape, especially in the context of document intelligence. Misalignment between these two stages can hinder a model's efficacy in subsequent tasks, as the model might find it challenging to apply its pre-acquired knowledge to the specialized requirements of fine-tuning tasks.

Recent studies spotlight two predominant chasms: discrepancies in the training objectives and variances in the knowledge needed for distinct tasks, as pointed out by~\cite{luo_geolayoutlm_2023}. The former relates to the incongruities in goals and architectures across the pre-training and fine-tuning phases. The latter pertains to the specific insights needed for a target task which might not be sufficiently grasped during pre-training. These challenges have birthed innovative methodologies. For example, Hu et al.~\cite{hu2022p3} aimed to bridge the objective discrepancy by aligning the goal of a subsequent ranking task with a pre-training one. Their findings revealed that such a harmonized training approach fostered more effective knowledge transfer.

Moreover, the emergence of prompt-based models has made the transition from pre-training to fine-tuning more seamless \cite{luo_geolayoutlm_2023}. By shaping downstream tasks into compatible prompts consistent with the pre-training objectives, these models enhance coherence across both stages. This results in increased adaptability and sets the stage for enhanced generalization for an array of tasks.

In the domain of document layout analysis, \cite{luo_geolayoutlm_2023} shed light on a formidable challenge when employing transformer-based models for RE tasks. The complex task of amalgamating textual, visual, and layout cues emphasized a marked objective misalignment during the RE's two phases. To counteract this, certain geometric-centric tasks were introduced to endow the model with pivotal geometric insights, aiming at better representation generalization harnessed from extensive pre-training datasets.

The emphasis on geometric comprehension has proven invaluable. Prior endeavours, such as StrucText \cite{li2021structext}, integrated geometric orientation details of textual segments in the pre-training phase. Yet, these primarily centred on exploring dyadic geometric relations. The avant-garde approach of extending these relations to encompass multi-pair and triplet configurations \cite{luo_geolayoutlm_2023} offers a more comprehensive perspective of document layouts.

To encapsulate, establishing a seamless bridge between the pre-training and fine-tuning stages is paramount for the success of models in diverse document analysis endeavours. As contemporary research showcases, refinements in training paradigms, task-oriented prompts, and an amplified emphasis on geometric insights can yield models endowed with resilience and adaptability across various document analysis challenges.

\section{Proposed Method}
\label{Proposed Method}
The proposed method involves a deep learning-based model that utilizes transformer-based models to encode all the information in a document image. This model is pre-trained and subsequently fine-tuned for various document image analysis tasks.

\subsection{Proposed Model Architecture}

The architecture of the proposed model is influenced by the LayoutLMv3 and GeoLayoutLM models, with alterations made to better accommodate the tasks at hand. The architecture is designed to encode all the information in a document image, including textual, visual, and layout information.

\begin{figure*}[ht]
    \centering
    \includegraphics[width=0.9\textwidth]{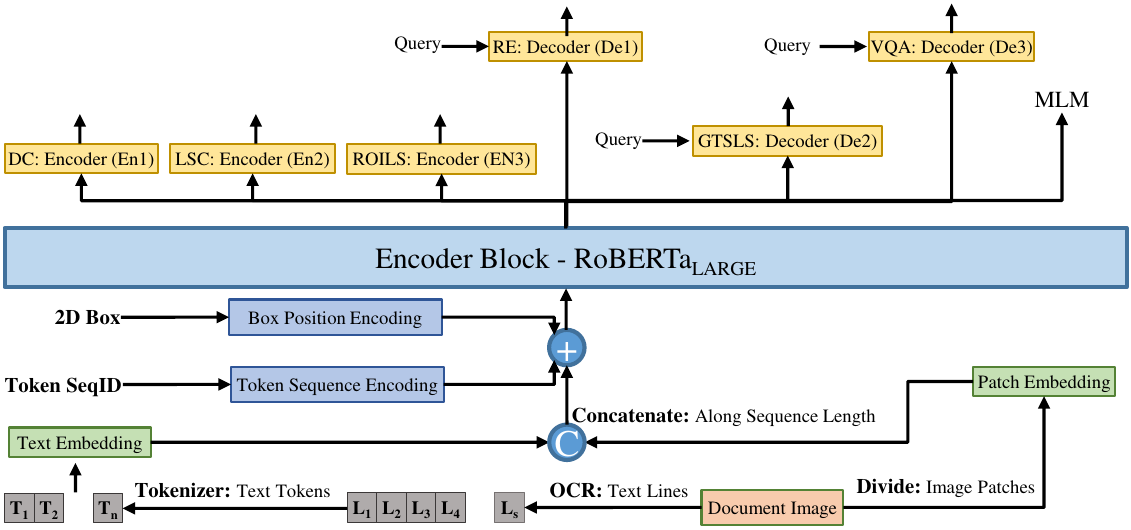}
\caption{A schematic of the suggested model showcasing various encoder and decoder units. Initially, text lines and their bounding boxes are identified via OCR. These lines undergo tokenization using a RoBERTa-based tokenizer. Subsequently, embeddings are produced for each token. Simultaneously, the input image is segmented into $32 \times 32$ patches, flattened and converted into patch embedding vectors. Both text and patch embeddings are combined with position encodings. The resulting vector sequence is then processed by the RoBERTa framework, passing through the designated encoder or decoder units, culminating in the final outputs.}
    \label{fig:Overview}
\end{figure*}

\subsubsection{Overview of the Proposed Model}

The proposed model is a deep learning-based model that integrates additional tasks during the pre-training phase. These tasks include identifying the reading order of different layout segments in a document image, categorizing layout segments as PubLayNet, and generating the text sequence within a given layout segment (text block). The model also incorporates a collective pre-training scheme where losses of all the tasks under consideration, including pre-training and fine-tuning tasks with all datasets, are considered. Additional encoder and decoder blocks are added to the RoBERTa network to generate results for all tasks~(refere the Figure~\ref{fig:Overview} for visual insights.).

\subsubsection{Detailed Model Architecture}

The model architecture consists of an independent text-embedding layer, patch embedding layer, position encoding, and multi-head attention (MHA) based encoder and decoder layers. The input of the proposed model is the document images and the text-line level OCR toolkit output. The input is first processed by corresponding text and patch embedding layers and converted into a latent feature vector. This vector is further added with respective encoding and then processed by the sequence of MHA layers.

\subsubsection{Text Embedding}
The OCR generates the text-line (Note: text with sufficient gap are considered different text-lines) level output with bounding box coordinates (coordinates are rescaled such that the longer side of the document image becomes 512). Each text line is first tokenized into a sequence of tokens, and then for each token, a latent feature vector is created by the text embedding layer (same as RoBERTa); we call these vectors a text latent feature vector and represented as $V_T$.

\subsubsection{Patch Embedding}
The document image is first resized such that the longer side of the document image becomes 512. The resized image is then divided into non-overlapping patches of size $32\times 32$. These patches are converted into vectors by flattening them. Then, a linear transformation (linear dense layer) is applied to convert them into latent feature vectors; we call these vectors visual latent feature vectors and represent them as $V_V$.

\subsubsection{Position Embedding}
There are two position embeddings: 1) Segment box embedding and 2) token sequence embedding within a segment box.

A segment's box is represented by $BOX_{seg} = x_1,y_1,x_2,y_2$ where $x_1,y_1$ are the coordinates of the left-top corner point of the box whereas the $x_2,y_2$ is the right-bottom corner point. The coordinates $BOX_{seg} = x_1,y_1,x_2,y_2 \in [1,512]$ thus a box (0,0,0,0) is not a valid box and utilized only when box coordinates are masked or not available.
The 1D encoding vectors for the $x_1,y_1,x_2, and y_2$ values are concatenated to form the segment box position vector. We use $P_B$ symbol to represent this vector.

A token sequence id is represented by $T_{seqid} \in [1,512]$; thus, a $T_{seqid}=0$ is not a valid token sequence id and is utilized only when there is no sequence available. The 1D encoding vector for the $T_{seqid}$ is used as the token sequence position vector. We use $P_t$ symbol to represent this vector.

The $P_B$ and $P_t$ position vectors are added together to form the final position vector, which is further added to the latent vectors $V_T$ or $V_V$. The resultant vectors are then processed by the sequence of MHA layers.

\subsubsection{Addition of Encoder and Decoder Blocks to RoBERTa Network}
Additional encoder and decoder blocks are added to the RoBERTa network to generate results for all tasks. The pre-training of the proposed model utilizes all these encoder and decoder blocks and updates the corresponding parameters as their respective losses. Different blocks are added according to different tasks; refer to section~\ref{Pre-training Tasks} "Pre-training Tasks" for better insights. 

\subsection{Pre-training Tasks}
\label{Pre-training Tasks}

The proposed model incorporates several pre-training tasks to enhance its performance. These tasks include:

\subsubsection{Masked Language Modeling (MLM)}
All datasets are used for this task in the pre-training phase of the model parameter learning. The masked language modelling inspires the MLM task in BERT~\cite{kenton2019bert}. A percentage of text tokens are masked with a span masking strategy, and the pre-training objective is to maximize the log-likelihood of the correct masked text tokens based on the contextual representations of corrupted sequences of image tokens and text tokens.

\subsubsection{Document Categorization (DC)}
The RVL-CDIP dataset~\cite{harley2015evaluation} is employed for the document image classification task. We have added a single MHA (cross attention) encoder layer (En1) with [CLS] token as the query vector with $BOX_{seg}= 0,0,0,0$ and $T_{seqid}=0$. Further, a linear transformation layer is added, followed by softmax activation for the document categorization.

\subsubsection{Layout Segment Categorization (LSC)}
The PubLayNet dataset~\cite{zhong2019publaynet} is used for the layout segment categorization. We have added a single MHA (cross attention) encoder layer (En2) with  [CLS] token as the query vector with $BOX_{seg}= layout~segment (x_1,y_1,x_2,y_2)$, $T_{seqid}=0$. Further, a linear transformation layer is added, followed by softmax activation for the document categorization.

\subsubsection{Reading Order Identification of Layout Segments (ROILS)}
The PubLayNet dataset~\cite{zhong2019publaynet} is used for the reading order identification of a given set of layout segments. We have added a single MHA (cross attention) encoder layer (En3) with a sequence of segment boxes as the query vector(each segment is encoded as $BOX_{seg}= layout~segment (x_1,y_1,x_2,y_2)$, $T_{seqid}=0$). This layer generates the output vector corresponding to each layout segment. These output vectors are further used in their reading order identification as introduced in the ERNIE\cite{peng2022ernie} model. This task helps the model to understand the flow of information in a document.

\subsubsection{Relation Extraction (RE)}
There are 4 different datasets, FUNSD~\cite{jaume2019funsd}, CORD~\cite{park2019cord}, SROIE~\cite{SROIE}, Kleister-NDA~\cite{Kleister-NDA},  used for this task. The relation extraction task involves identifying the relationships between different entities in the document. The output sequence length is also not fixed; therefore, we employed a decoder block for this task.

We have added a block of two MHA (casual and cross attention) decoder layers (De1) with the description text of the required relation as query (each token of this text has $BOX_{seg}=(0,0,0,0)$). This block generates the next text token until the [EOS] token is generated.

\subsubsection{Generation of Text Sequence within Layout Segments (GTSLS)}
The PubLayNet dataset~\cite{zhong2019publaynet} is used for this task. Due to the same reason, this task also requires a decoder block. Therefore, We have added a block of two MHA (casual and cross attention) decoder layers~(De2) with [SOS] token as the query vector with $BOX_{seg}= layout~segment (x_1,y_1,x_2,y_2)$, $T_{seqid}=0$. This block generates the next text token until the [EOS] token is generated.

\subsubsection{Visual Question Answering (VQA)}
In the visual question-answering task, the model is trained to generate the answer for a question according to the information available inside the document image. This task helps the model understand the document's content in a more detailed and comprehensive manner. The DocVQA~\cite{mathew2021docvqa} dataset is used for this task. We have added a block of three MHA (casual and cross attention) decoder layers~(De3) with the question text as query (each token of this text has $BOX_{seg}=(0,0,0,0)$). This block generates the next text token until the [EOS] token is generated.

\section{Experimental Setup}
\label{Experimental Setup}
It is important to note that the experimental setup, which includes the datasets and evaluation metrics, is in line with the most recent models, such as LayoutLMv3 and GeoLayoutLM~\cite{huang2022layoutlmv3,luo_geolayoutlm_2023}. This makes sure that our experiments are measured against the most recent progress in the field. We are utilising 4 GPUs, V100, for the pre-training and fine-tuning the proposed model.

\begin{table}[!hbt]
\centering
\caption{Statistics of datasets}
\label{table:dataset_statistics}
\begin{tabular}{c|c|c}
\toprule
\multirow{2}{2cm}{\centering \textbf{Dataset} }& \multirow{2}{2cm}{\centering \textbf{\# of keys or categories} } & \multirow{2}{2cm}{\centering \textbf{\# of examples (train/dev/test)} }\\
& & \\
\midrule
RVL-CDIP~\cite{harley2015evaluation} & 16 & 320K/4K/4K \\
FUNSD~\cite{jaume2019funsd} & 4 & 149/0/50 \\
CORD~\cite{park2019cord} & 30 & 800/100/100 \\
SROIE~\cite{SROIE} & 4 & 626/0/347 \\
Kleister-NDA~\cite{Kleister-NDA} & 4 & 254/83/203 \\
DocVQA~\cite{mathew2021docvqa} & – & 39K/5K/5K \\
PubLayNet~\cite{zhong2019publaynet}  & 5 & 335703/11245/11405\\
\bottomrule
\end{tabular}
\end{table}

\subsection{Datasets}
The statistics of different datasets used in all of our experiments regarding pre-training and fine-tuning of the proposed model is listed in Table~\ref{table:dataset_statistics}. The descriptions of different datasets according to their targeted tasks are as follows:

\subsubsection{\textbf{Dataset for Layout Segment Analysis}}
The PubLayNet dataset~\cite{zhong2019publaynet} stands out as a substantial resource for research on document layout analysis, mainly when using deep learning approaches. The dataset encompasses a comprehensive collection of research paper images, each meticulously annotated with bounding boxes and polygonal segmentations. The main objective behind this annotation is to distinctly categorise various segments of the document layout, which have been broadly divided into five categories: text, title, list, figure, and table.

The dataset's structure and organisation have been designed to facilitate both training and evaluation processes. The official partitioning of the dataset includes a training set comprising 335,703 images, a validation set with 11,245 images, and a test set with 11,405 images. The division ensures that the models can be thoroughly trained on vast data and then validated and tested for performance consistency.

Moreover, the inclusion of different layout structures in PubLayNet aligns well with the proposed work's objective of reading order identification, layout segment categorisation, and generating the text sequence within specific layout segments. Utilising this dataset for pre-training tasks, as intended in the proposed model, could bridge the objective gap observed between the pre-training and fine-tuning phases.

\begin{figure*}[ht]
    \centering
    \includegraphics[width=0.9\textwidth]{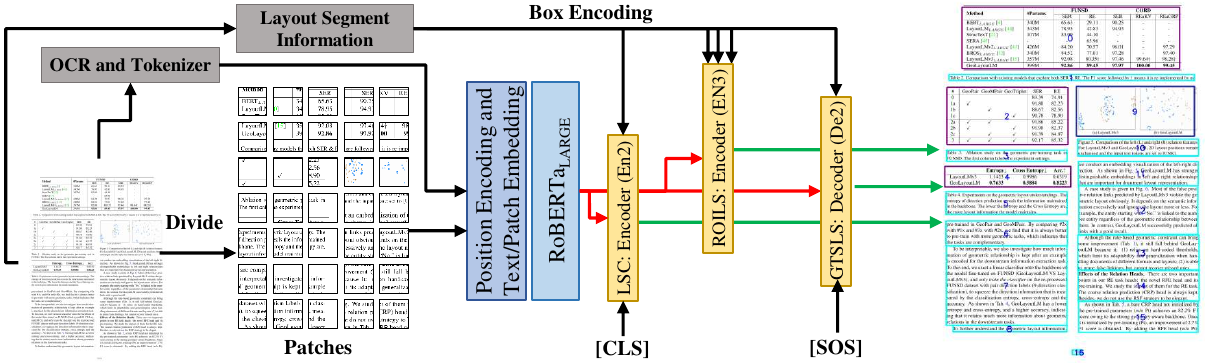}
\caption{Illustration of utilising the PubLayNet dataset for pre-training tasks in the proposed model. The dataset, containing a rich collection of document layout images, is employed to pre-train the model parameters for subsequent layout segment analysis tasks. The figure depicts how the dataset facilitates reading order identification, layout segment categorisation, and text sequence generation within specific layout segments, aligning with the objectives of the proposed work.}

    \label{fig:PubLayNet}
\end{figure*}

\textbf{Note:} This dataset is utilised only for the pre-training phase of the model parameter learning as we do not have the required components to get the layout segment bounding box from the given document image (an object detection network or a segmentation-based network). A pictorial depiction of the utilisation of this dataset in different pre-training tasks is given in Figure~\ref{fig:PubLayNet}.

\subsubsection{\textbf{Dataset for Document Classification}}
The RVL-CDIP dataset~\cite{harley2015evaluation} is employed for the document image classification task. This dataset is a subset of the IIT-CDIP collection, containing document images labelled with 16 distinct categories. The RVL-CDIP dataset comprises 400,000 document images distributed into 320,000 training images, 40,000 validation images, and 40,000 test images. Text and layout information are extracted using the Microsoft Read API~\cite{huang2022layoutlmv3}

\subsubsection{\textbf{Dataset for Entity Relation Extraction}}
For entity relation extraction, we used two primary datasets:

\textit{\textbf{FUNSD Dataset~\cite{jaume2019funsd}:}} The FUNSD dataset is a collection of noisy scanned forms for document understanding and analysis. It consists of 199 documents that provide comprehensive annotations for 9,707 semantic entities. The objective is to label each semantic entity as \enquote{question}, \enquote{answer}, \enquote{header}, or \enquote{other}. The dataset is split into 149 training and 50 test samples.

\textit{\textbf{CORD Dataset~\cite{park2019cord}:}} CORD is dedicated to the extraction of key information from receipts. It is segmented into 800 training, 100 validation, and 100 test samples and houses 30 semantic labels distributed under four categories.

\textit{\textbf{SROIE Dataset~\cite{SROIE}:}} The SROIE dataset, unveiled by Huang et al. during the ICDAR2019 Competition for Scanned Receipt OCR and Information Extraction, encompasses 1000 entire scanned receipt images along with annotations. The dataset is divided into a training set of 626 documents and a testing set of 347 documents, categorised under four distinct classes. It was crafted for a contest centred on OCR (Optical Character Recognition) and the extraction of crucial information from scanned receipts. The SROIE dataset is a valuable asset for the progress, assessment, and refinement of OCR and key information extraction methodologies, particularly in scanned receipts.

\textit{\textbf{Kleister-NDA~\cite{Kleister-NDA}}} 
The \enquote{Kleister-NDA} dataset, tailored for Key Information Extraction (KIE) tasks, includes a blend of scanned and born-digital formal English documents, mainly Non-disclosure Agreements (NDAs). It houses 540 NDAs (training set: 254, validation set: 83, test set: 20), encompassing 3,229 unique pages and 2,160 entities for extraction. Initially sourced from the Electronic Data Gathering, Analysis and Retrieval system (EDGAR) via Google, these documents were converted to PDFs and annotated for entity extraction through a two-tier process involving three annotators and a super-annotator.

\subsubsection{\textbf{Dataset for Document Visual Question Answering}}
For the document visual question answering (DocVQA) task, the DocVQA dataset~\cite{mathew2021docvqa} is employed. This dataset is designed for visual question answering over document images and includes a partition of 10,194 training images, 1,286 validation images, and 1,287 test images. Additionally, there are 39,463 questions for the training set, 5,349 for the validation set, and 5,188 for the test set. The objective is to input a document image alongside a question and expect the model to produce an accurate answer.

\subsection{Evaluation Metrics}
\subsubsection{\textbf{Metrics for Document Classification}}
The main evaluation metric used for document classification is accuracy, which measures the proportion of correctly classified documents against the total documents.

\subsubsection{\textbf{Metrics for Entity Relation Extraction}}
The F1 score is employed as the primary metric for the entity relation extraction task. The F1 score provides a harmonic mean of precision (ratio of correctly predicted positive observations to the total predicted positives) and recall (ratio of correctly predicted positive observations to all observations in actual class). It is especially useful when the dataset has imbalanced classes.

\subsubsection{\textbf{Metrics for Document Visual Question Answering}}
For document visual question answering, the commonly-used edit distance-based metric ANLS (also known as Average Normalized Levenshtein Similarity) is reported.

\subsection{Model Pre-training}
The proposed model leverages the rich information in collecting datasets for different document analysis tasks. The collection utilised in the proposed work is consist of the dataset as RVL-CDIP~\cite{harley2015evaluation}, FUNSD~\cite{jaume2019funsd}, CORD~\cite{park2019cord}, SROIE~\cite{SROIE}, Kleister-NDA~\cite{Kleister-NDA}, DocVQA~\cite{mathew2021docvqa}, and PubLayNet~\cite{zhong2019publaynet}. The collective training set of all these datasets is one million (1M). Building upon the foundational works of LayoutLMv3~\cite{huang2022layoutlmv3} and GeoLayoutLM~\cite{luo_geolayoutlm_2023}. The matched layers of the proposed model are initialised from the weightings of RoBERTa~\cite{liu_roberta_2019} and the remaining are initialised from random distribution, same as LayoutLMv3~\cite{huang2022layoutlmv3}. This hybrid approach seamlessly merges textual and visual understanding, providing a more holistic representation of documents. The Adam optimiser is used to pre-train the model with a sizeable batch size of 32 over 2,500,000 steps. We have samples from all datasets for each batch as RVL-CDIP: 8, FUNSD: 2, CORD: 2, SROIE: 2, Kleister-NDA: 2, DocVQA: 8, PubLayNet: 8 samples. 

\subsection{Model Fine-tuning}
Each task necessitates a unique fine-tuning approach:
\begin{itemize}
\item \textbf{Document Classification:} Based on the experiments conducted on the RVL-CDIP~\cite{harley2015evaluation} dataset, the model is adjusted over 50,000 steps. This fine-tuning employs a batch size of 32 with a learning rate of \(2\times10^{-5}\).

\item \textbf{Entity Relation Extraction:} Four datasets,  FUNSD~\cite{jaume2019funsd}, CORD~\cite{park2019cord}, SROIE~\cite{SROIE}, Kleister-NDA~\cite{Kleister-NDA}, are considered for this task. The model undergoes a fine-tuning with batch size 32, epochs 100, and the learning rate is adjusted to $3\times10^{-5}$ for each dataset separately.

\item \textbf{Document Visual Question Answering:} This task, based on the DocVQA dataset, requires the model to interpret both a document image and a related question, aiming to output a relevant answer. The base model is adjusted over 500,000 steps using a batch size of 32, a learning rate of $2\times10^{-5}$, and a warmup ratio of 0.05.
\end{itemize}

\section{Result and Analysis}
\label{Result and Analysis}
This section presents the results obtained from the proposed model and provides a detailed analysis. The results are categorized based on the tasks performed by the model. 

\subsection{Analysis of Document Classification Performance}

\setlength{\tabcolsep}{2pt}

\begin{table*}[!hbt]
\centering
\caption{Comparative classification accuracies on the RVL-CDIP dataset. Note: "R/G/P" implies the "region/grid/patch" image embedding types.}
\begin{tabular}{c|c|c|c|c|c|c}
\toprule
\textbf{Modality} & \textbf{Model} & \textbf{Image Embedding} & \multicolumn{2}{|c|}{\textbf{Pre-Training}} & \textbf{Accuracy} & \textbf{\#Parameters} \\
& & & \textbf{Dataset Size} & \textbf{Epochs} & & \\
\midrule
\multirow{4}{3cm}{\centering Text only} & BERT$_{BASE}$ & None & - & - & 89.81\% & 110M \\
& RoBERTa$_{BASE}$ & None & - & - & 90.06\% & 125M \\
& BERT$_{LARGE}$ & None & - & - & 89.92\% & 340M \\
& RoBERTa$_{LARGE}$ & None & - & - & 90.11\% & 355M \\
\midrule
\multirow{7}{*}{Text + Layout + Image} & DocFormer$_{BASE}$~\cite{appalaraju2021docformer} & (G) ResNet-50 & - & - & 96.17\% & 183M \\
& LayoutLMv1$_{BASE}$\cite{xu2020layoutlm} & (R) ResNet-101 & 11M & 2 & 94.42\% & 160M \\
& LayoutLMv2$_{BASE}$~\cite{xu2021layoutlmv2} & (G) ResNeXt101-FPN & 11M & 5 & 95.25\% & 200M \\
& LayoutLMv3$_{BASE}$\cite{huang2022layoutlmv3} & (P) Linear & 11M & 90 & 95.44\% & 133M \\
& DocFormer$_{LARGE}$~\cite{appalaraju2021docformer} & (G) ResNet-50 & - & - & 95.50\% & 536M \\
& LayoutLMv2$_{LARGE}$~\cite{xu2021layoutlmv2} & (G) ResNeXt101-FPN & 11M & 20 & 95.25\% & 426M \\
& LayoutLMv3$_{LARGE}$~\cite{huang2022layoutlmv3} & (P) Linear & 11M & 90 & 95.44\% & 368M \\
\midrule
\multirow{2}{*}{Text + Layout + Image} & Our Proposed Model & \multirow{2}{*}{(P) Linear} & \multirow{2}{*}{1M} & \multirow{2}{*}{80} & \multirow{2}{*}{\textbf{95.87}\%} & \multirow{2}{*}{390M} \\
&  BackBone network RoBERTa$_{LARGE}$ & & & & \\
\bottomrule
\end{tabular}
\label{tab:Document Classification Results on RVL-CDIP}
\end{table*}
In analyzing the results from our experiments on the RVL-CDIP dataset (Table~\ref{tab:Document Classification Results on RVL-CDIP}), the proposed model registered an impressive accuracy of 95.87\%. This achievement places our model at a competitive standing amidst leading-edge models in the document classification domain.

Several observations and takeaways can be gleaned from these outcomes:
\begin{itemize}
\item \textbf{The Power of Hybrid Training:} One of the most distinguishing features of our model's success was its ability to harness information from textual, visual, and layout sources. It is evident from Table~\ref{tab:Document Classification Results on RVL-CDIP} that models leveraging a combination of these elements consistently outperformed text-only models.

\item \textbf{Intrinsic Value of Enhanced Pre-training:} The meticulous incorporation of specialized pre-training tasks, such as reading order identification and layout segment categorization, arguably fortified the model's capability to discern document layouts and their inherent content. This intrinsic understanding was particularly pivotal in accurately classifying complex documents with sophisticated layouts.

\item \textbf{Efficiency in Parameter Utilization:} Despite having a parameter count of 390M, which is considerably fewer than some counterparts, our model was able to maintain competitive performance. This speaks to its efficiency and optimization in parameter utilization.

\item \textbf{Comparative Analysis:} In direct comparison with the widely recognized LayoutLM series and DocFormer models, our proposed model stands out due to its superior understanding of document layout and content and its unique pre-training schemes and architecture adjustments.

\item \textbf{Image Embedding Insights:} The models that employ diverse image embeddings showcase varied performance. For instance, the "patch" type of image embedding used in our proposed model and LayoutLMv3 seems to strike a harmonious balance between accuracy and model complexity.
\end{itemize}
In conclusion, the advancements we integrated into the model's architecture and training regimen have borne fruit in the form of enhanced classification accuracy. This reinforces the significance of tailored pre-training tasks and architectural refinements in deep learning-based document analysis endeavours.

\subsection{Analysis and Interpretation of Entity Relation Extraction Results}

\begin{table*}[!hbt]
\centering
\caption{Entity-level F1 scores of entity extraction tasks on the four datasets: FUNSD, CORD, SROIE and Kleister-NDA. \textbf{Note:} "R/G/P" implies the "region/grid/patch" image embedding types.}
\begin{tabular}{c|c|c|cccc|c}
\toprule

\textbf{Modality} & \textbf{Model} & \textbf{Image Embedding} & \textbf{FUNSD} & \textbf{CORD} & \textbf{SROIE} & \textbf{Kleister-NDA} & \textbf{\#Parameters} \\
\midrule
\multirow{4}{3cm}{\centering Text only} & BERT$_{BASE}$ & None & 0.6026 & 0.8968 & 0.9099 & 0.7790 & 110M\\
 & UniLMv2$_{BASE}$~\cite{bao2020unilmv2} & None & 0.6648 & 0.9092 & 0.9459 & 0.7950 & 125M\\
 & BERT$_{LARGE}$ & None & 0.6563 & 0.9025 & 0.9200 & 0.7910 & 340M\\
 & UniLMv2$_{LARGE}$~\cite{bao2020unilmv2} & None & 0.7072 & 0.9205 & 0.9488 & 0.8180 & 355M\\
\midrule
 \multirow{2}{*}{Text + Layout} & BROS$_{BASE}$~~\cite{hong2022bros} & None & 0.8305 & 0.9573 & 0.9548 & - & 110M \\
 & BROS$_{LARGE}$~~\cite{hong2022bros} & None & 0.8452 & 0.9740 & - & - & 340M \\
\midrule
 \multirow{4}{*}{Text + Layout + Image} & DocFormer$_{BASE}$~\cite{appalaraju2021docformer} & (G) ResNet-50 & 0.8334 & 0.9633 & - & - & 183M \\
& LayoutLMv1$_{BASE}$\cite{xu2020layoutlm} & (R) ResNet-101 & 0.7866 & 0.9472 & 0.9438 & 0.8270 & 160M\\
& LayoutLMv2$_{BASE}$~\cite{xu2021layoutlmv2} & (G) ResNeXt101-FPN & 0.8276 & 0.9495 & 0.9625 & 0.8330 & 200M \\
& LayoutLMv3$_{BASE}$\cite{huang2022layoutlmv3} & (P) Linear & 0.9029 & 0.9656 & - & - & 133M \\
& DocFormer$_{LARGE}$~\cite{appalaraju2021docformer} & (G) ResNet-50 & 0.8455 & 0.9699 & - & - & 536M \\
& LayoutLMv1$_{LARGE}$\cite{xu2020layoutlm} & (R) ResNet-101 & 0.7895 & 0.9493 & 0.9524 & 0.8340 & 343M \\
& LayoutLMv2$_{LARGE}$~\cite{xu2021layoutlmv2} & (G) ResNeXt101-FPN & 0.8420 & 0.9601 & 0.9781 & 0.8520 & 426M \\
& LayoutLMv3$_{LARGE}$~\cite{huang2022layoutlmv3} & (P) Linear & 0.9208 & 0.9746 & - & - & 368M \\
& GeoLayoutLM~\cite{luo_geolayoutlm_2023} & (R) ConvNeXt-FPN & 0.9286 & 0.9797 & - & - & 399M \\
\midrule
\multirow{2}{*}{Text + Layout + Image} & Our Proposed Model & \multirow{2}{*}{(P) Linear} & \multirow{2}{*}{\textbf{0.9306}} & \multirow{2}{*}{\textbf{0.9804}} & \multirow{2}{*}{\textbf{0.9794}} & \multirow{2}{*}{\textbf{0.8742}} & \multirow{2}{*}{425M} \\
&  BackBone network RoBERTa$_{LARGE}$ & & & & \\
\bottomrule
\end{tabular}
\label{tab:Entity-level F1}
\end{table*}

The proposed model was subjected to rigorous testing on four distinct datasets for the entity relation extraction task, namely FUNSD, CORD, SROIE, and Kleister-NDA. As depicted in Table~\ref{tab:Entity-level F1}, the model achieved impressive F1 scores across all datasets, with 0.9306 on FUNSD, 0.9804 on CORD, 0.9794 on SROIE, and 0.8742 on Kleister-NDA. 

These high F1 scores are indicative of the model's superior performance in accurately extracting relations between entities in complex documents. This can be attributed to the model's comprehensive understanding of the document structure and content, which directly results from incorporating three additional tasks during the pre-training phase. These tasks include reading order identification of different layout segments in a document image, layout segments categorization as per PubLayNet, and generation of the text sequence within a given layout segment (text block).

The proposed model outperforms other models in the same category, such as GeoLayoutLM~\cite{luo_geolayoutlm_2023} and LayoutLMv3$_{LARGE}$~\cite{huang2022layoutlmv3}, in terms of F1 scores on the FUNSD and CORD datasets. This suggests that the proposed model is more effective in handling complex documents with intricate layouts and multiple entities. 

On the SROIE dataset, the proposed model achieved a comparable F1 score to LayoutLMv2$_{LARGE}$~\cite{xu2021layoutlmv2}, indicating its effectiveness in extracting key information from scanned receipts. 

On the Kleister-NDA dataset, the proposed model achieved an F1 score of 0.8742, significantly higher than other models in the same category. This demonstrates the model's ability to accurately extract entities from formal English documents, such as Non-disclosure Agreements (NDAs).

In conclusion, the proposed model demonstrates superior performance across various documents in entity relation extraction tasks. The high F1 scores achieved on all datasets validate the effectiveness of the additional tasks incorporated during the pre-training phase. The model's ability to understand and interpret complex document layouts and content makes it a promising tool for document analysis tasks.

\subsection{Document Visual Question Answering Results}

\begin{table}[!hbt]
\centering
\caption{ANLS score on the DocVQA dataset (PM: Parameters)}
\begin{tabular}{c||c|c|c}
\toprule
\textbf{Modality} & \textbf{Model} & \textbf{ANSL} & \textbf{\#PM} \\
\midrule
\multirow{4}{3cm}{\centering Text only}
 & BERT$_{BASE}$ & 0.6354 & 110M\\
 & UniLMv2$_{BASE}$~\cite{bao2020unilmv2} & 0.7134 & 125M\\
 & BERT$_{LARGE}$ & 0.6768 & 340M\\
 & UniLMv2$_{LARGE}$~\cite{bao2020unilmv2} & 0.7709 & 355M\\
\midrule
\multirow{6}{*}{Text + Layout + Image} 
& LayoutLMv1$_{BASE}$\cite{xu2020layoutlm} &  0.6979 & 160M\\
& LayoutLMv2$_{BASE}$~\cite{xu2021layoutlmv2} & 0.7808 & 200M \\
& LayoutLMv3$_{BASE}$\cite{huang2022layoutlmv3} & 0.7876 & 133M \\
& LayoutLMv1$_{LARGE}$\cite{xu2020layoutlm} & 0.7259 & 343M \\
& LayoutLMv2$_{LARGE}$~\cite{xu2021layoutlmv2} & 0.8348 & 426M \\
& LayoutLMv3$_{LARGE}$~\cite{huang2022layoutlmv3} & 0.8337 & 368M \\
\midrule
\multirow{2}{*}{Text + Layout + Image} & Our Proposed Model & \multirow{2}{*}{\textbf{0.8468}} & \multirow{2}{*}{440M} \\
& BackBone RoBERTa$_{LARGE}$ &  & \\
\bottomrule
\end{tabular}
\label{tab:ANLS score on the DocVQA dataset}
\end{table}

The Document Visual Question Answering (DocVQA) task demands a delicate blend of visual, layout, and textual comprehension. Our experimentation reveals significant insights into the models' performance on this multifaceted task. Purely text-based models, such as BERT and UniLMv2, tend to have restricted performance capabilities, as evident from their ANLS scores in Table~\ref{tab:ANLS score on the DocVQA dataset}. While BERT, even in its large configuration, couldn't cross the 0.7 mark, UniLMv2$_{LARGE}$ managed to achieve a score of 0.7709. The pronounced improvement in ANLS scores when transitioning from text-only models to those incorporating layout and image information is striking. LayoutLM versions, especially, show significant strides in their performance. By leveraging the additional modalities, these models are better positioned to comprehend the complexities of document images holistically.

Our proposed model, with additional encoder-decoder layers, achieves an impressive ANLS score of 0.8468. Not only does this score surpass the benchmarks set by the other models, but it also underscores the effectiveness of our approach. The collective pre-training scheme adopted, integrating diverse datasets and tasks, is pivotal in refining our model's understanding. By leveraging this diverse knowledge base, our model excels in extracting intricate details from documents, making it proficient at answering nuanced queries. The results highlight the fundamental importance of multi-modality integration in the DocVQA task. The fusion of textual, visual, and layout information, combined with innovative pre-training strategies, has the potential to push the boundaries of what models can achieve in document analysis.

\subsection{Ablation Study: Analysis of Pre-training Tasks}

\begin{table*}[!hbt]
\centering
\caption{Ablation Study of the proposed model with different Pre-training Tasks. The performance of different datasets with respect to their corresponding task is listed here. The Document Classification task is evaluated as accuracy (\%), the Relation Extraction task as F1 score, and the Visual Question Answering task as the ANSL score.}
\begin{tabular}{c|c|cccc|c}
\toprule

\multirow{2}{3cm}{\centering \textbf{Task}}  &  \textbf{Document Classification} & \multicolumn{4}{|c|}{\textbf{Relation Extraction}} & \textbf{Visual Question Answering} \\
 & \textbf{RVL-CDIP} & \textbf{FUNSD} & \textbf{CORD} & \textbf{SROIE} & \textbf{Kleister-NDA} & \textbf{DocVQA} \\
\midrule
 MLM + DC + LSC & 95.49\% & 0.8343 & 0.9452 & 0.9529 & 0.8530 & 0.7959\\
\midrule
MLM + DC + LSC + RE + VQA & 95.38\% & 0.8843 & 0.9552 & 0.9629 & 0.8603 & 0.8334\\
\midrule
\centering MLM + DC + LSC + RE + VQA + ROILS + GTSLS & \textbf{95.87}\% & \textbf{0.9306} & \textbf{0.9804} & \textbf{0.9794} & \textbf{0.8742} & \textbf{0.8468}\\
\bottomrule
\end{tabular}
\label{tab:Ablation Study}
\end{table*}
In the light of recent breakthroughs in document analysis, particularly the cutting-edge models presented in~\cite{huang2022layoutlmv3, luo_geolayoutlm_2023}, a noticeable performance improvement is observed for the $BERT_{LARGE}$ and $RoBERTa_{LARGE}$ models compared to their base versions, $BERT_{BASE}$ and $RoBERTa_{BASE}$. Our study focuses on the $RoBERTa_{LARGE}$ model, intertwined with various pre-training tasks, aiming to unravel and clarify these tasks' individual and combined contributions.

Our analysis reveals several key insights:

\textbf{Reading Order Identification of Layout Segments~(ROILS):} This task is a cornerstone for the model's performance across all tasks. Its significant impact suggests that understanding the spatial sequence of text in a document is crucial. It serves as a roadmap for models to decode the logical flow, inherently enhancing contextual comprehension.

\textbf{Layout Segments Categorization (LSC):} The inclusion of LSC led to a considerable improvement in Document Classification and Entity Relation Extraction tasks. This highlights the importance of identifying different layouts within a document, enabling a clearer delineation of segments and their hierarchical relationships.

\textbf{Generation of Text Sequence within Layout Segments (GTSLS):} This task plays a crucial role in enhancing the Document Visual Question Answering task. By allowing the model to generate textual sequences within specific layouts, it equips the model with a more detailed understanding of the content, facilitating more accurate information retrieval. However, its impact on the Entity Relation Extraction task was less pronounced, indicating potential areas for further improvement.

In conclusion, while each pre-training task contributed to enhancing the model's performance to varying extents, their collective inclusion resulted in the most significant improvements. This underscores that a comprehensive understanding of a document, with its complex interplay of textual, visual, and spatial information, requires a multifaceted approach.


\bibliography{citation}
\bibliographystyle{IEEEtran}

\end{document}